%% file: main.tex
\documentclass[conference,a4paper]{IEEEtran}
\usepackage{multirow}
\usepackage[pdftex]{graphicx}
\usepackage{amsmath}
\usepackage[psamsfonts]{amssymb}
\usepackage{amsxtra}
\usepackage{threeparttable}
\usepackage{url}
\usepackage{bm}
\usepackage{algorithm}
\usepackage{algorithmic}
\usepackage{comment}

\graphicspath{{}}
\newcommand{\figref}[1]{Fig. \ref{figure:#1}}
\newcommand{\tabref}[1]{Table \ref{table:#1}}
\newcommand{\eqnref}[1]{Eq. \ref{eq:#1}}

\begin{document}

\title{Sensing and Navigation of Aerial Robot for \\ Measuring Tree Location and Size \\ in Forest Environment}

\author{%
\IEEEauthorblockN{%
Tomoki Anzai\IEEEauthorrefmark{1},
Moju Zhao\IEEEauthorrefmark{1},
Fan Shi\IEEEauthorrefmark{1},
Kei Okada\IEEEauthorrefmark{1},
Masayuki Inaba\IEEEauthorrefmark{1}
}
\IEEEauthorblockA{%
\IEEEauthorrefmark{1}%
The University of Tokyo, Japan}
}

\maketitle
\thispagestyle{empty}

\begin{abstract}
\input src/abst.tex
\end{abstract}

\input src/introduction.tex
\input src/tree_recognition.tex
\input src/tree_database.tex

\input src/searching_method.tex
\input src/experiment.tex
\newpage
\input src/conclusion.tex

\bibliographystyle{unsrt}
\bibliography{main}

\end{document}

%% file: src/abst.tex
This paper shows the achievement of a sensing and navigation system of aerial robot for measuring location and size of trees in a forest environment autonomously.  Although forestry is an important industry in Japan, the working population of forestry is decreasing. Then, as an application of mechanization of forestry, we propose tree data collection system by aerial robots which have high mobility in three-dimensional space. First, we develop tree recognition and measurement method, along with algorithm to generate tree database. Second, we describe aerial robot navigation system based on tree recognition. Finally, we present an experimental result in which an aerial robot flies in a forest and collects tree data.

%% file: src/introduction.tex
\section{Introduction}
In Japan, forestry is an important industry because approximately seventy percent of the area of the land is forest. However, nowadays forestry faces decline and the working population decreases for some reasons. One of the reasons is that the accident rate of forestry is much higher than other industries. To solve this problem, mechanization of forestry has been promoted \cite{Attebrant1997}\cite{Westerberg2014}.
\par
Recently, aerial robots have attracted a lot of attention and have been studied actively due to their high mobility in three-dimensional environments \cite{Kumar2012}\cite{Lindsey2012}. Particularly, aerial vision, such as surveillance\cite{Doitsidis2012}, vision based SLAM(Simultaneous Localization and Mapping)\cite{Milford2011}\cite{Mueggler2014}, etc. is an active research area. 
\par
\begin{figure}[t]
\begin{center}
\includegraphics[bb=0 0 606 241, width=1.0\linewidth]{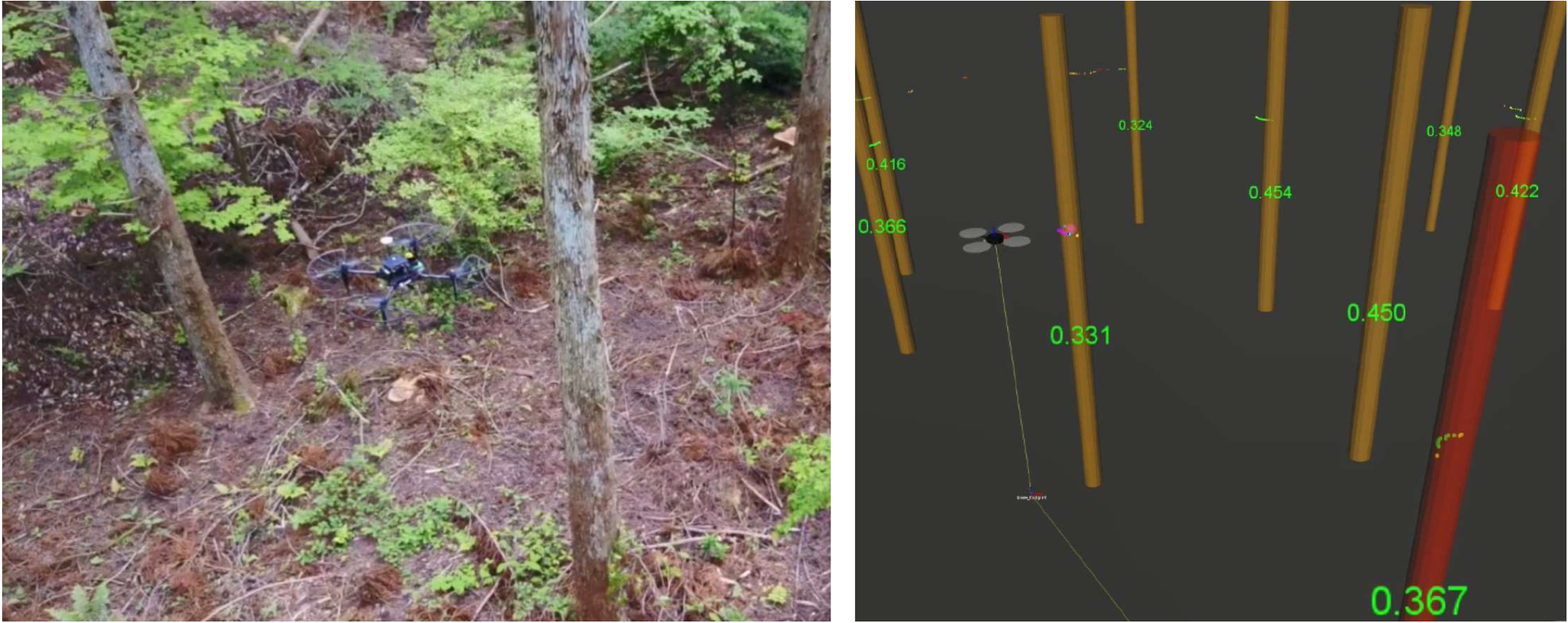}
\end{center}
\caption{Left: an aerial robot flying in a forest environment to collect data of tree location and size. Right: tree location and size data obtained by aerial robot with proposed method.
\label{figure:forest_drone}}
\end{figure}

As an application of aerial vision, we propose a system in which an aerial robot flies in the air automatically and collects data of tree location and size(\figref{forest_drone}). In forest, we can not enter some areas, for instance, steep hill, swamp, etc. However, aerial robots can fly over such areas and collect tree data. We consider this system will be helpful for forestry. To achieve this system, tree detection and measurement method is necessary for collecting data. To generate forest environment data, we must develop tree database generating method. The navigation system of aerial robot is also necessary for automatic flight.
\par
In this paper, the main purpose is to achieve the sensing and navigation system of aerial robot for measuring forest environment. In Sec. II, we develop tree recognition and measurement method with LRF and camera. In Sec. III, we propose an algorithm to generate tree database. Sec. IV describes aerial robot navigation system based on tree recognition. Finally, we present an experimental result in Sec. V to demonstrate the feasibility of proposed tree measurement system.

%% file: src/tree_recognition.tex
\section{Tree recognition and measurement method}
\begin{figure}[t]
\begin{center}
\includegraphics[bb=0 0 417 240, width=0.9\linewidth]{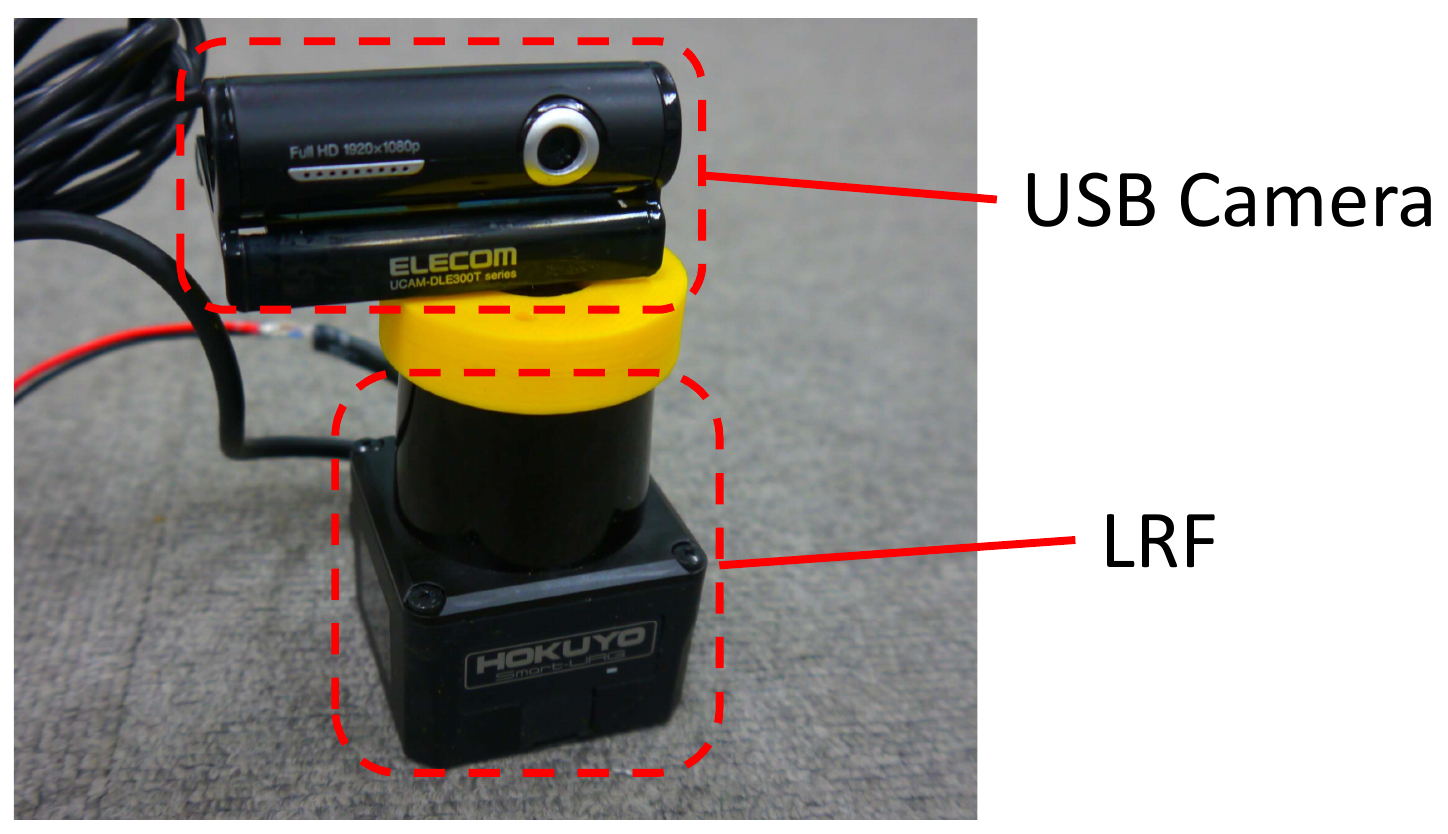}
\end{center}
\caption{The forest sensing device. The components are as follows: \newline
1) Hokuyo UST-20LX-H01: 2D Laser Range Finder. The maximum range is 20[m] and the scan angle is 270[deg]. \newline
2) ELECOM UCAM-DLE300TBK: USB camera. The horizontal and vertical angle of view are 54[deg] and 42[deg].
\label{figure:sensor_device}}
\end{figure}

\figref{sensor_device} shows the device we use for sensing forest environment. The device consists of an LRF and a USB camera. The yellow part made with 3D printer is a mount for the USB camera. The LRF is Hokuyo UST-20LX-H01 which can measure the distance to objects within the range of 20[m] and the scan angle is 270[deg]. The USB camera is ELECOM UCAM-DLE300TBK: the horizontal and vertical angle of view are 54[deg] and 42[deg], respectively. The center of LRF is matched with the center of image of USB camera.
We can use Hokuyo LRF and USB camera easily in ROS(Robot Operating System) since the drivers\cite{urgnode}\cite{usbcamera} are developed. In this section, we describe how to detect and measure trees by the range information from LRF and image from USB camera.

\subsection{Laser Scan Data Clustering}
The range data has a lot of noise when we scan a real environment with LRF. Therefore, noise filtering is necessary to obtain precise information. 
\par
First, to remove invalid range values which can occur when the measured object is too near or too far from the LRF, we apply range filter. The function of range filter can be written as:

\begin{equation}
	\begin{cases}
  \text{\it{Valid}} \hspace{0.1\columnwidth} (\text{\it{Thre}}_L < range < \text{\it{Thre}}_H)\\
  \text{\it{Invalid}} \hspace{0.07\columnwidth} (otherwise)
  \end{cases}
  \label{eq:range_filter}
\end{equation}

where $\text{\it{Thre}}_L$ and $\text{\it{Thre}}_H$ are lower threshold and upper threshold, respectively. 
\par
Second, to remove discontinuous points which are caused by veiling effect, we apply shadow filter. The function of shadow filter can be written as:

\begin{equation}
	\begin{cases}
  \text{\it{Valid}} \hspace{0.1\columnwidth} (\theta_{min}  < \theta < \theta_{max})\\
  \text{\it{Invalid}} \hspace{0.07\columnwidth} (otherwise)
  \end{cases}
  \label{eq:shadow_filter}
\end{equation}

\begin{equation}
	\theta = \angle O P_i P_{i+1}
\end{equation}

where $O$, $P_i$ and $P_{i+1}$ are the origin of the LRF, a scanned point and the next scanned point, respectively. If $\theta$ is near to 0[deg] or 180[deg], $P_i$ and $P_{i+1}$ can be considered as discontinuous points. 
\par
Finally, we perform clustering considering valid continuous points as a cluster since shadow filter filters out discontinuous points.

\subsection{Tree Measurement and Discrimination}
\subsubsection{Tree Size and Location Measuring}
\begin{figure}[t]
\begin{center}
\includegraphics[bb=0 0 469 265, width=1.0\linewidth]{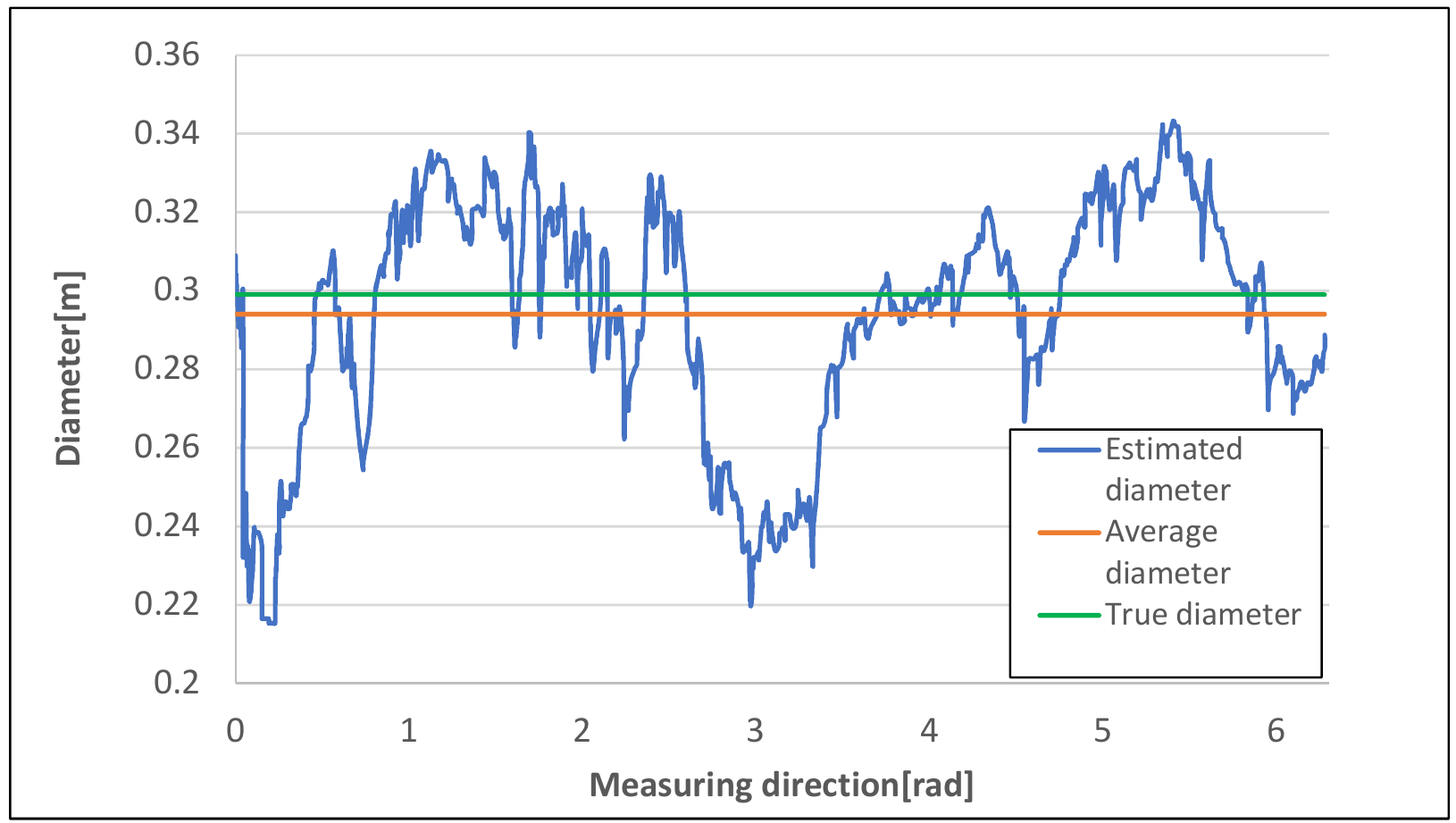}
\end{center}
\caption{Result of circle fitting of a real tree. The true diameter is 0.299[m] and the average estimated diameter is 0.294[m]. \label{figure:fitting_result}}
\end{figure}

If a cluster obtained from range data is a tree, we can extract the information of tree size and location approximating a tree as cylinder object. In this work, we approximate shape of trees as a cylinder. We use least-square method to perform circle fitting. Given that scanned points are $P_i(x_i, y_i)$, this is written as:

\begin{equation}
	\begin{bmatrix}
	A \\
	B \\
	C \\
	\end{bmatrix}
  =
  \begin{bmatrix}
    \sum_i x_i^2 &\sum_i x_iy_i &\sum_i x_i\\
    \sum_i x_iy_i &\sum_i y_i^2 &\sum_i y_i \\
    \sum_i x_i &\sum_i y_i &\sum_i 1
  \end{bmatrix}
	^{-1}
  \begin{bmatrix}
    -\sum_i (x_i^3+x_iy_i^2) \\
    -\sum_i (x_i^2y_i+y_i^3) \\
    -\sum_i (x_i^2+y_i^2)
  \end{bmatrix}
  \label{eq:least_square_method}
\end{equation}
\begin{equation}
	A = -2a; \ B = -2b; \ C = a^2+b^2-r^2 
\end{equation}

where $(a, b)$ and $r$ are the center position and the radius of the estimated circle, respectively. \figref{fitting_result} shows the experimental result of measuring tree. In this experiment, an aerial robot with LRF goes around a tree. The true diameter is calculated by dividing measured circumference by $\pi$. Since the tree is not a complete cylinder, the diameter changes with the direction of measuring and the maximum error is 0.082[m]. However, the difference between the true diameter(0.299[m]) and the average diameter(0.294[m]) is 0.005[m]. Therefore, it is assumed that the averaged diameter is reliable. 

\subsubsection{Tree Discrimination}
Each cluster obtained from range data might be a tree or not. Therefore, each cluster must be judged as a tree or not by its feature quantity. 
\par
First, we consider the residual sum of squares which can be calculated as follows:

\begin{equation}
	S = \sum_i \{ (x_i - a)^2 + (y_i - b)^2 - r^2\} ^2 
	\label{eq:residual}
\end{equation}

The least-square method(\eqnref{least_square_method}) outputs the solution which minimizes $S$( \eqnref{residual}). If the points $P_i(x_i, y_i)$ are on a complete circle, $S$ equals to $0$. To the contrary, if $S$ is large, the probability of being a tree is low. Note that $S$ scales linearly with $r^4$ and the number of points. To take this into account, we must use coefficient of variation($CV$) to evaluate the probability. This is written as:

\begin{equation}
	CV = \sqrt{\frac{S}{Nr^4}}
	\label{eq:residual}
\end{equation}

where $N$ is the number of points in the cluster.

\par
\begin{figure}[t]
\begin{center}
\includegraphics[bb=0 0 711 410, width=1.0\linewidth]{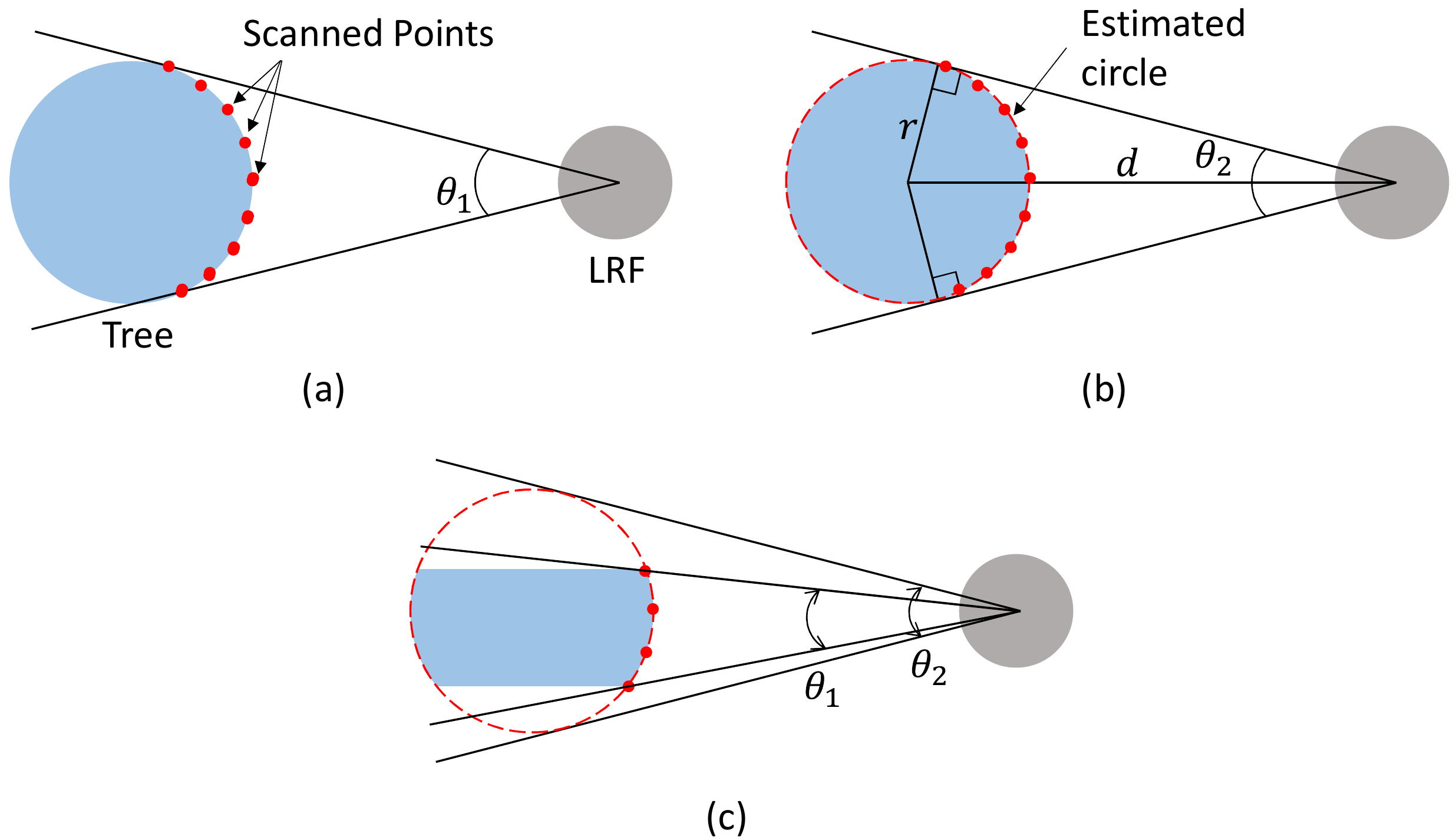}
\end{center}
\caption{(a): Real angle of view($\theta_1$). (b): Imaginary angle of view($\theta_2$) which equals to $\theta_1$ if the shape of the object is cylinder. (c): A case in which the shape of the scanned object is not cylinder. $\theta_1$ differs from $\theta_2$. \label{figure:scan_angle}}
\end{figure}

Second, we consider the angle of view of objects. $\theta_1$ in \figref{scan_angle}(a) is the real angle of view where $\theta_2$ in \figref{scan_angle}(b) is the imaginary angle of view. They can be calculated as:

\begin{eqnarray}
	\theta_1 &=& N \cdot \theta_{step} \\
	\theta_2 &=& 2\arcsin \frac{d}{r}; \ d = \sqrt{a^2+b^2}
\end{eqnarray}
 
Note that $\theta_{step}$ is the angular distance between two adjacent measured points which is a constant parameter of an LRF. $\theta_{step}$ of Hokuyo UST-20LX-H01 is 0.25[deg]. If the cluster is a cylinder, $\theta_1$ and $\theta_2$ is approximately the same value. To the contrary, as shown in \figref{scan_angle}(c), the difference between $\theta_1$ and $\theta_2$ is caused if the scanned object is not a cylinder. 
\par
Finally, if the tree radius estimated by \eqnref{least_square_method} is too small or large, the possibility of the cluster being a tree is low.
\par
To sum up, these discriminating methods described above can be written as:
\begin{equation}
  \begin{split}
  \begin{cases}
  \text{\it{Valid}} \hspace{0.1\columnwidth} (CV < \text{\it{Thre}}_{CV} \ \text{\it{AND}} \\
  \hspace{0.2\columnwidth}  |\theta_1 - \theta_2| < \text{\it{Thre}}_{\theta_{view}} \ \text{\it{AND}} \\  
  \hspace{0.2\columnwidth}  r_{min} < r < r_{max})\\
  \text{\it{Invalid}} \hspace{0.07\columnwidth} (otherwise)
  \end{cases}
  \end{split}
  \label{eq:discrimination}
\end{equation}
 
\subsection{Labeled Tree Detection}
\begin{figure}[t]
\begin{center}
\includegraphics[bb=0 0 615 419, width=1.0\linewidth]{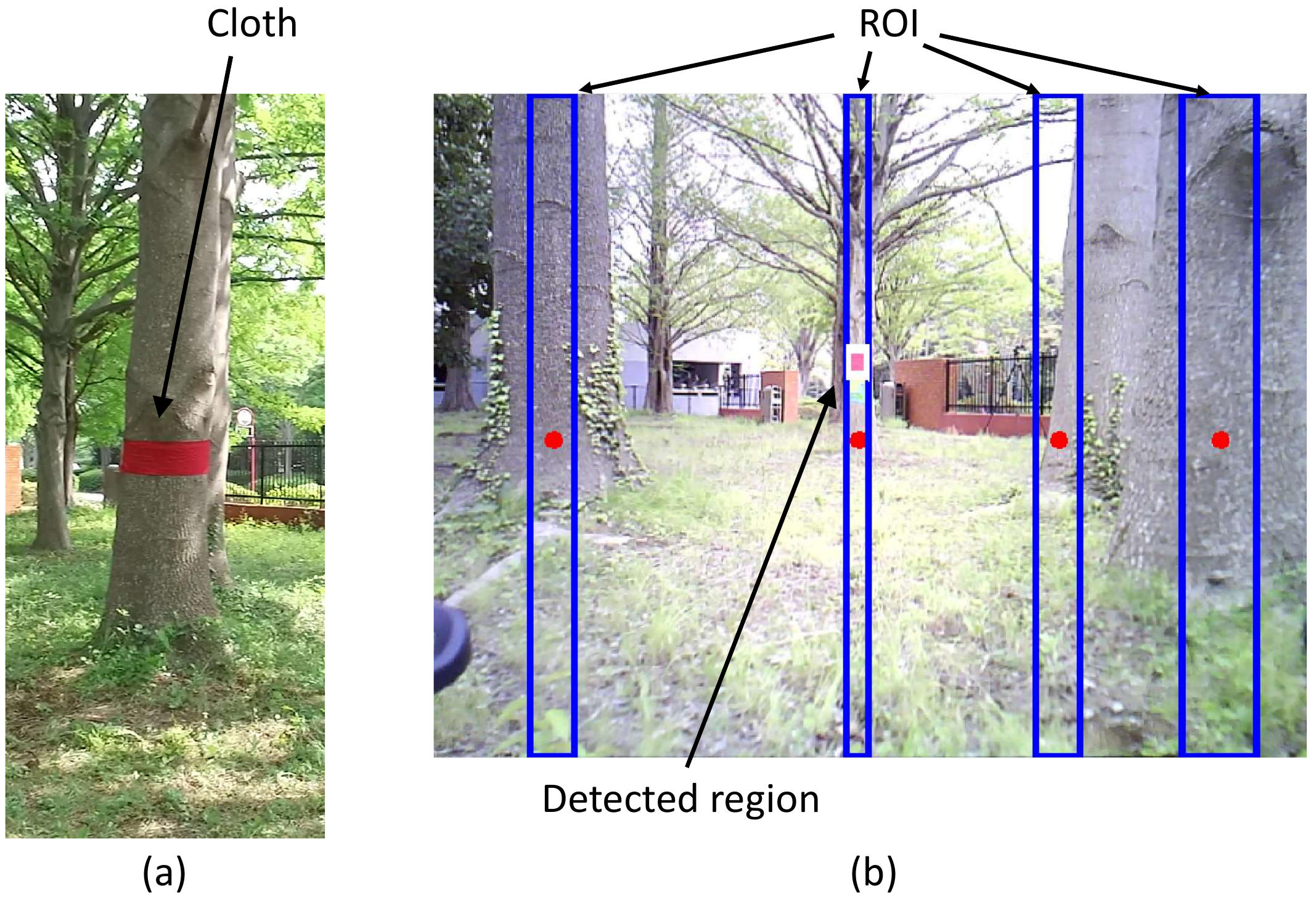}
\end{center}
\caption{(a): Labeled tree with red cloth. (b): Labeled tree detection method: the blue rectangular frames are ROI and white contour in one of the ROIs is detected region. \label{figure:labeled_tree}}
\end{figure}

In this work, we label a tree with red cloth as shown in \figref{labeled_tree}(a). We can compare two data of trees in a forest easily by identifying the labeled tree in two data. In this section, we explain how to detect the labeled tree. Since the center of LRF is matched with the center of image of USB camera, ROI(Region of Interest) in the image can be determined by tree size and position data obtained by the clustering and discriminating method. In \figref{labeled_tree}(b), the blue rectangular frames indicate the ROI. Next, to detect the red cloth, HSV color filter is applied for each ROI. In \figref{labeled_tree}(b), the region surrounded by white contour is the detected region. 

%% file: src/tree_database.tex
\section{Tree database and mapping}
We described the method to detect trees in forest environment in Sec. II. In this section, we explain how to make tree database and map of forest.
\par
In this work, we consider forest environment as two-dimensional space. Therefore, the parameter of each tree is two-dimensional position($x, y$) and radius($r$). From the range data of LRF, a tree position in LRF frame can be obtained. Given that LRF is fixed to a aerial robot and the frame of LRF and the base link of the aerial robot is the same, the tree position in world frame can be calculated as:
\begin{equation}
  \begin{bmatrix}
    x\\
    y
  \end{bmatrix}
	^{\{W\}} = 
	\begin{bmatrix}
    \cos \varphi_{uav}& -\sin \varphi_{uav} \\ 
    \sin \varphi_{uav}& \cos \varphi_{uav}
  \end{bmatrix}
	\begin{bmatrix}
    x\\ 
    y
  \end{bmatrix}
	^{\{LRF\}} 
	+
	\begin{bmatrix}
    x_{uav}\\
    y_{uav}
  \end{bmatrix}
	^{\{W\}} 	
\end{equation}

where $(x_{uav}, y_{uav})$ and $\varphi_{uav}$ is the position and Euler angle yaw of the aerial robot, respectively. Since the aerial robot is always around hovering state, the roll and pitch angles
can be regarded low. In this work, we obtain the position of the aerial robot from optical flow sensor and the yaw angle from IMU(Inertia Measurement Unit) and compass sensor. 
\par
In Algorithm. \ref{tree_database}, we show the tree database update algorithm. The return value of getNewTreeList() is the list of new tree data and the frame of the tree position is world frame. Then, to determine the nearest tree in the database, each distance between the new tree data and trees in database is calculated. If a tree which is enough near to the new tree exists, they can be regarded as the same tree, and its position and radius are updated. In addition, the tree data has votes cast as a parameter. When the tree data is updated, the votes cast increments. Thus, the reliability of being a tree can be evaluated by the votes cast. If there is no tree near to the new tree data in database, the new tree data can be regarded as a new tree, and then we add the new tree data to database as a new tree.
\par
Note that the odometry of aerial robot($x_{uav}, y_{uav}$) drifts during flight. However, it does not matter with respect to updating tree database because the tree positions in the world frame are compared in two continuous frames where the drift is sufficiently low.

\begin{algorithm}                      
\caption{Tree database update algorithm \label{tree_database}}                                
\begin{algorithmic}                  
\STATE {$L \leftarrow$ getNewTreeList()}
\FORALL {$new\_tree$ in $L$}
	\STATE {$nearest\_tree \leftarrow tree\_db$.getNearestTree($new\_tree$)} 
	\IF {$nearest\_tree$.distanceTo($new\_tree) < Thre_{dist}$} 
		\STATE {$nearest\_tree$.updateData($new\_tree$)}
		\STATE {$nearest\_tree$.vote()}
	\ELSE
		\STATE {$tree\_db$.addNewTree($new\_tree$)}
	\ENDIF
\ENDFOR

\end{algorithmic}
\end{algorithm}
\par
In addition to tree database, using gmapping\cite{gmapping}, we make 2D map of forest. Gmapping requires laser scan and odometry information. 
\par
\begin{figure}[t]
\begin{center}
\includegraphics[bb=0 0 261 220, width=1.0\linewidth]{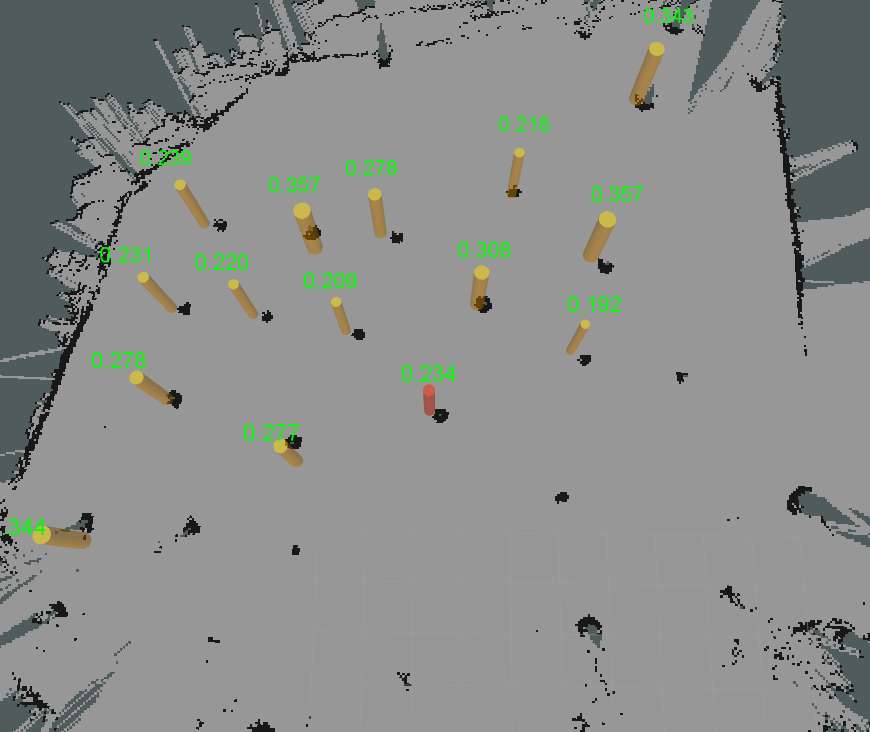}
\end{center}
\caption{The tree database and map generated in an experiment in a real forest. \label{figure:database}}
\end{figure}

In \figref{database}, the result of making tree database and map generated in a real forest is shown. The cylinder objects indicate the measured trees, and the red one located nearly at the center of the image is the labeled tree. The value above each tree is the estimated radius. The gray area is the generated map while the black areas indicate obstacles such as trees, net, etc. With comparing the position of trees indicated by cylinder and the black holes in the map, it is evident that the tree database and map generation are fully performed.

%% file: src/searching_method.tex
\section{Aerial Robot Navigation}
In this section, we describe aerial robot navigation method based on tree positions obtained with the methods described in Sec. II and Sec. III by an aerial robot with the sensing device(\figref{sensor_device}). In this work, we use standard quad-rotor which has four degree of freedom of motion($\phi, \theta, \varphi, z$). While there are some types of operation of quad-rotor, we use velocity control to generate aerial motion. The operating velocities are $(v_x, v_y, v_z, v_\varphi)$.
\subsection{Circular Motion}
\begin{figure}[t]
\begin{center}
\includegraphics[bb=0 0 444 216, width=1.0\linewidth]{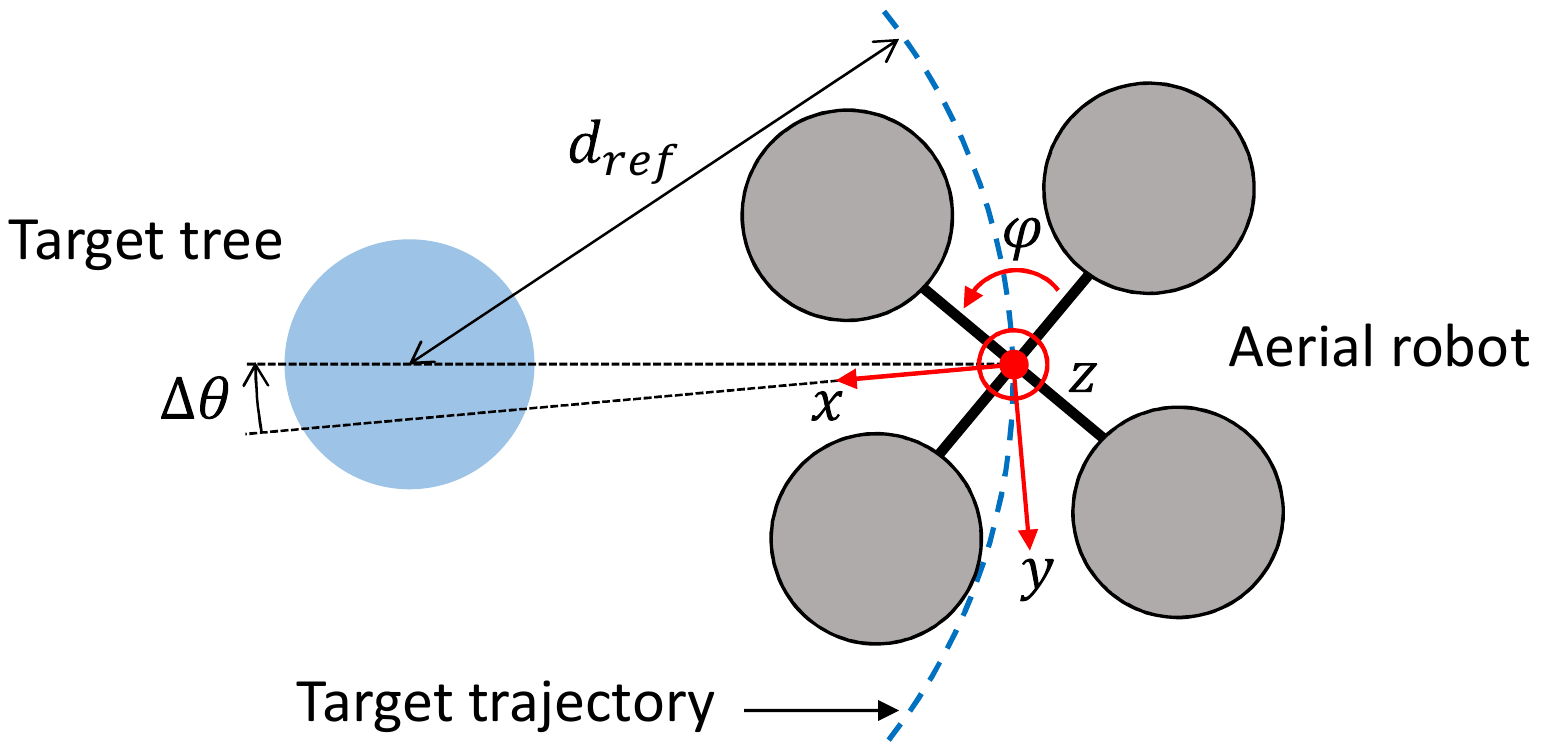}
\end{center}
\caption{Explanation of circular motion. The right-handed coordinate is set on the aerial robot.
\label{figure:circular_motion}}
\end{figure}

\begin{figure}[t]
\begin{center}
\includegraphics[bb=0 0 394 347, width=0.9\linewidth]{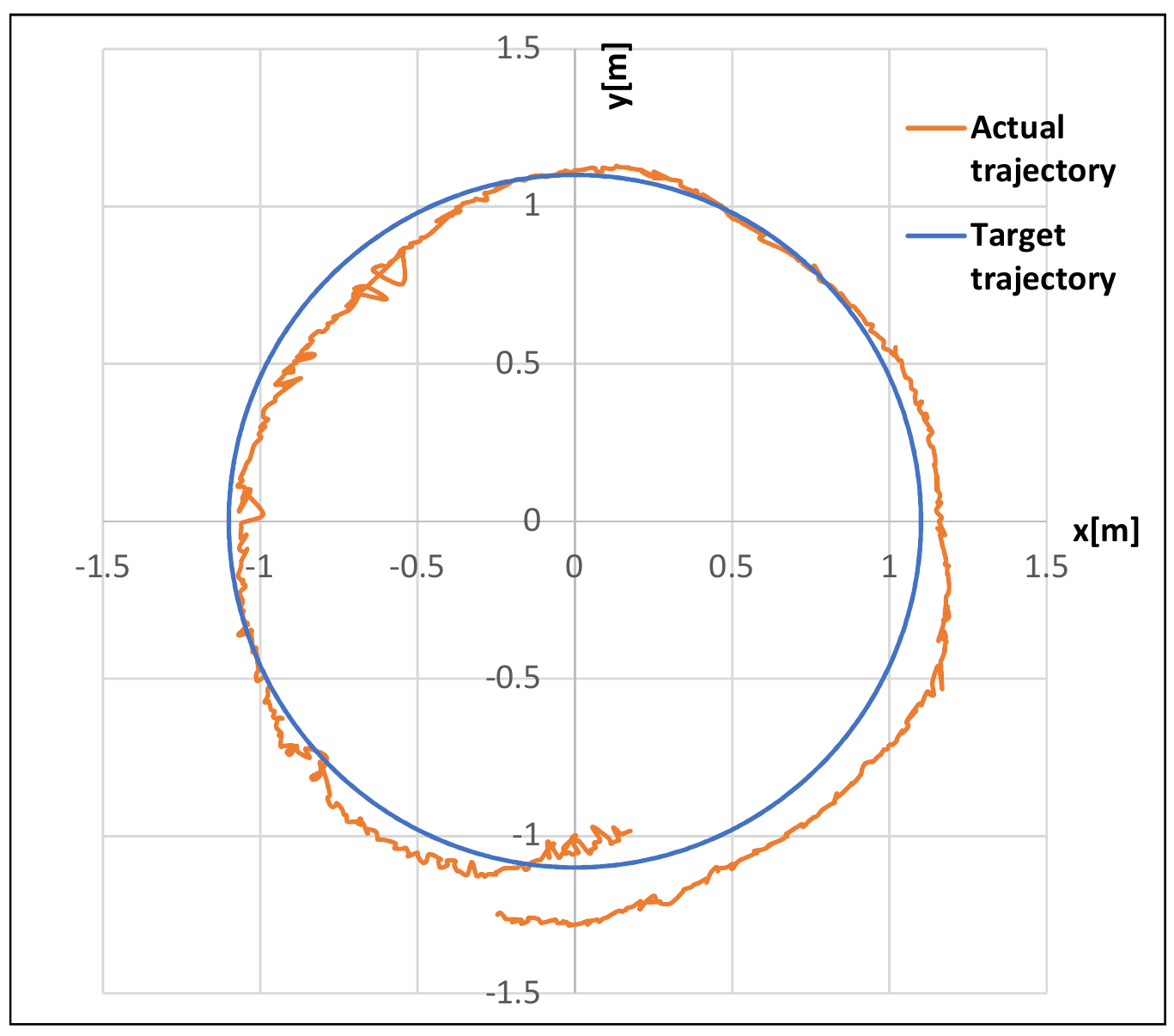}
\end{center}
\caption{Result of circular motion: the center of the tree is $(0, 0)$ and the actual and target trajectory of the aerial robot are shown.
\label{figure:circular_motion_result}}
\end{figure}

To measure a tree size precisely, the aerial robot goes around the tree. As shown in \figref{circular_motion}, we set the right-handed coordinate system on the aerial robot. In circular motion, we calculate velocity command as follows:

\begin{eqnarray}
	v_x &=& K_x (d_{ref} - d) \label{eq:x_control} \\
	v_y &=& V \\
	v_z &=& K_z (z_{ref} - z) \\
	v_\varphi &=& K_\varphi \Delta \theta - V / d_{ref} \label{eq:yaw_control}
\end{eqnarray}

where $V$ is a constant and $K_x$, $K_z$ and $K_\varphi$ are proportional gains.  The distance to the target tree is controlled by \eqnref{x_control}. In \eqnref{yaw_control}, the first term of the right side is feedback component and the second term is feedforward component. The experimental result of circular motion is shown in \figref{circular_motion_result}. $d_{ref}$ was set to 1.1[m], that is, the target trajectory is a circle whose center and radius are the center of the target tree and 1.1[m] respectively. Although the maximum error is about +0.2[m], it can be considered that circular motion is successfully generated. The cause of the error can be assumed to be centrifugal force.

\subsection{Searching Method in Forest Environment}
The searching algorithm we use is shown in Algorithm. \ref{searching_algorithm}. After the aerial robot goes around the target tree, the target tree will be changed to another tree which has not been measured. Then, we must consider the method of next tree determination, in other words, the implementation of getNextTargetTree() function. We implement two methods as follows:

\subsubsection{Narrow Searching Method}
In narrow searching method,  we set searching area shown in \figref{searching_methods}(a). The searching area is circle area whose center is the labeled tree. After the labeled tree is measured, trees in the searching area are selected one by one counterclockwise (or clockwise) as the next target tree. 
\subsubsection{Deep Searching Method}
\begin{figure}[t]
\begin{center}
\includegraphics[bb=0 0 680 285, width=1.0\linewidth]{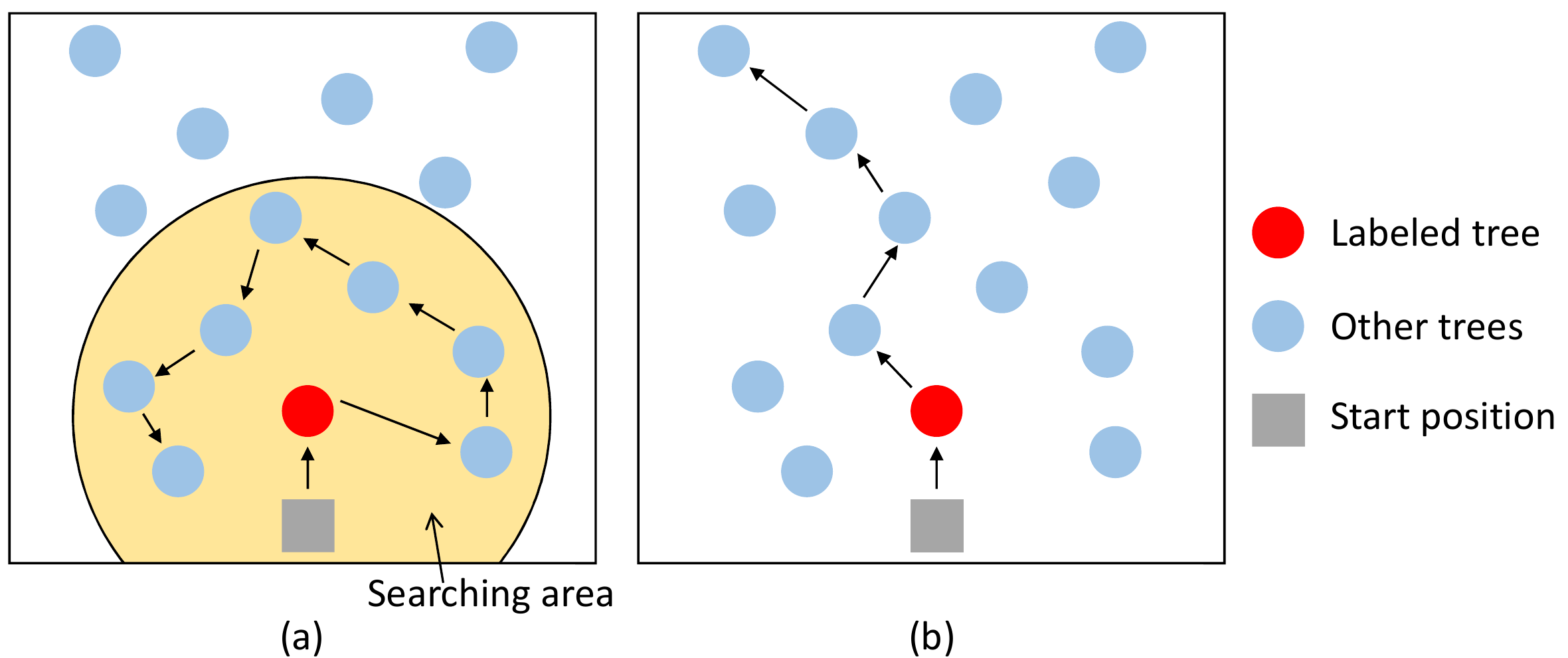}
\end{center}
\caption{(a): Narrow searching method. Trees in the searching area are selected counterclockwise. (b): Deep searching method. The tree which locates deeper than the previous target tree and nearest from the previous target tree is selected.
\label{figure:searching_methods}}
\end{figure}

In deep searching method, there is no limited searching area. As shown in \figref{searching_methods}(b), the tree which locates deeper than the previous target tree and nearest from the previous target tree is selected as the next target tree.

\begin{algorithm}                      
\caption{Searching algorithm \label{searching_algorithm}}                              
\begin{algorithmic}
\STATE{$target\_tree \leftarrow labeled\_tree$}            
\LOOP
	\STATE{$uav$.approachTo($target\_tree$)}
	\STATE{$uav$.circularMotion()}
	\STATE{$target\_tree \leftarrow tree\_db.$getNextTargetTree()}
	\IF {$target\_tree$ is $None$}
			\STATE{break}
	\ENDIF
\ENDLOOP
\end{algorithmic}
\end{algorithm}

%% file: src/experiment.tex
\section{Experiment}
\subsection{Aerial Robot Platform}
\begin{figure}[t]
\begin{center}
\includegraphics[bb=0 0 423 284, width=0.9\linewidth]{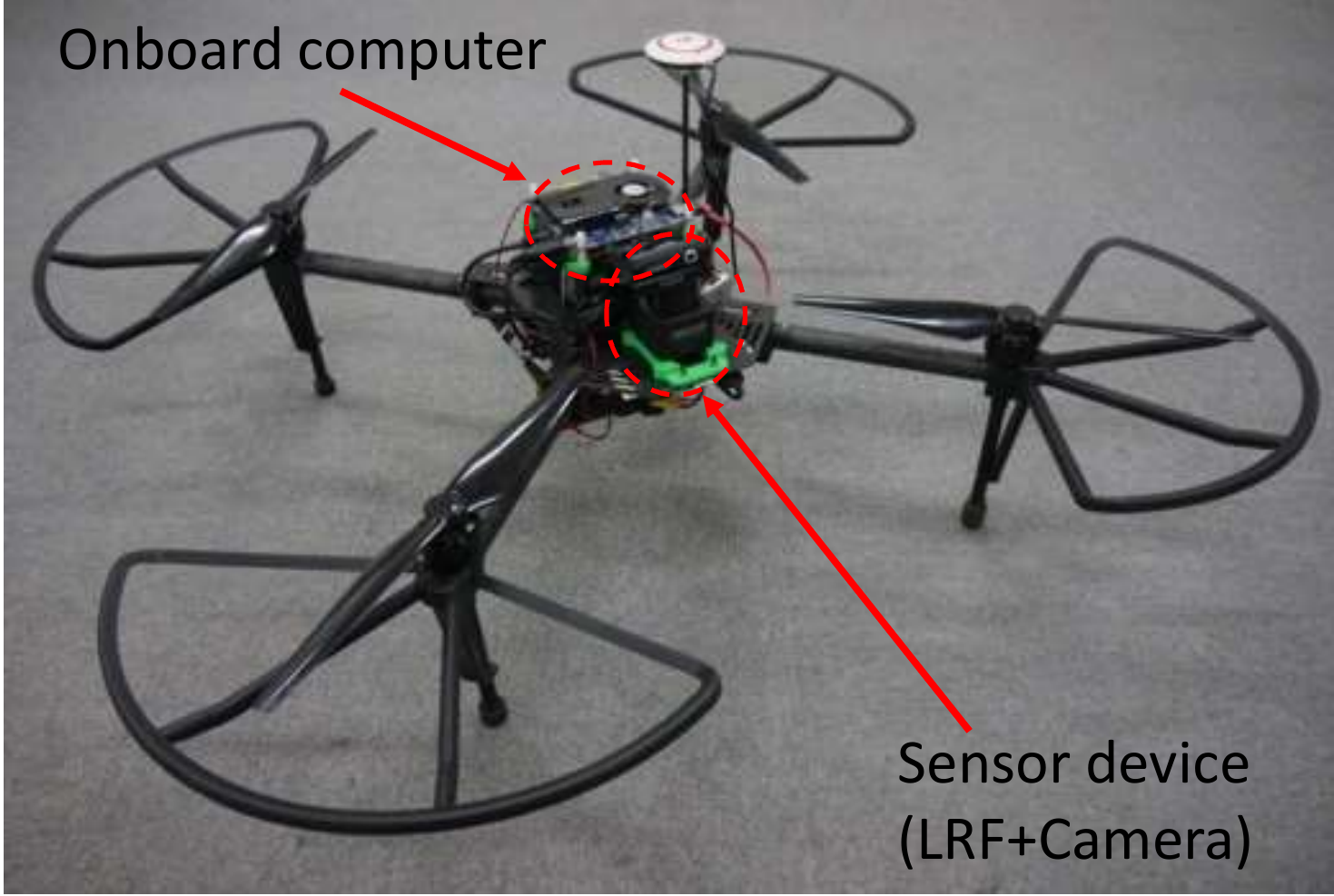}
\end{center}
\caption{DJI M100 with an onboard computer and the sensor device.
\label{figure:m100}}
\end{figure}

We conducted flight experiments using DJI M100 shown in \figref{m100}. M100 is a quad-rotor for development and programmable with DJI SDK\cite{djisdk}. Our M100 comprises an onboard computer and the sensor device introduced in Sec. II. The whole weight including battery is 2.35[kg] and hovering time is about 22[min].

\subsection{Tree Measurement Experiment in Forest Environment}
We performed tree measurement experiments in a real forest. We participated in forest drone robot competition held by the Center for Socio-Robotic Synthesis, Kyushu Institute of Technology in May, 2017\cite{forestdronechallenge}. The venue for the competition is in Yufuin, Oita. In the competition, there are three tasks. In the first task, the aerial robot approaches to the labeled tree within 1.0[m]. In the second task, the aerial robot goes around the labeled tree and measures the diameter of the labeled tree. In the third task, after the aerial robot goes around the labeled tree, it goes around other trees, and measures the diameter of trees. After the tasks, if we submit the data of diameter of the trees, we gain additional scores.
\par
\begin{figure}[t]
\begin{center}
\includegraphics[bb=0 0 314 517, width=1.0\linewidth]{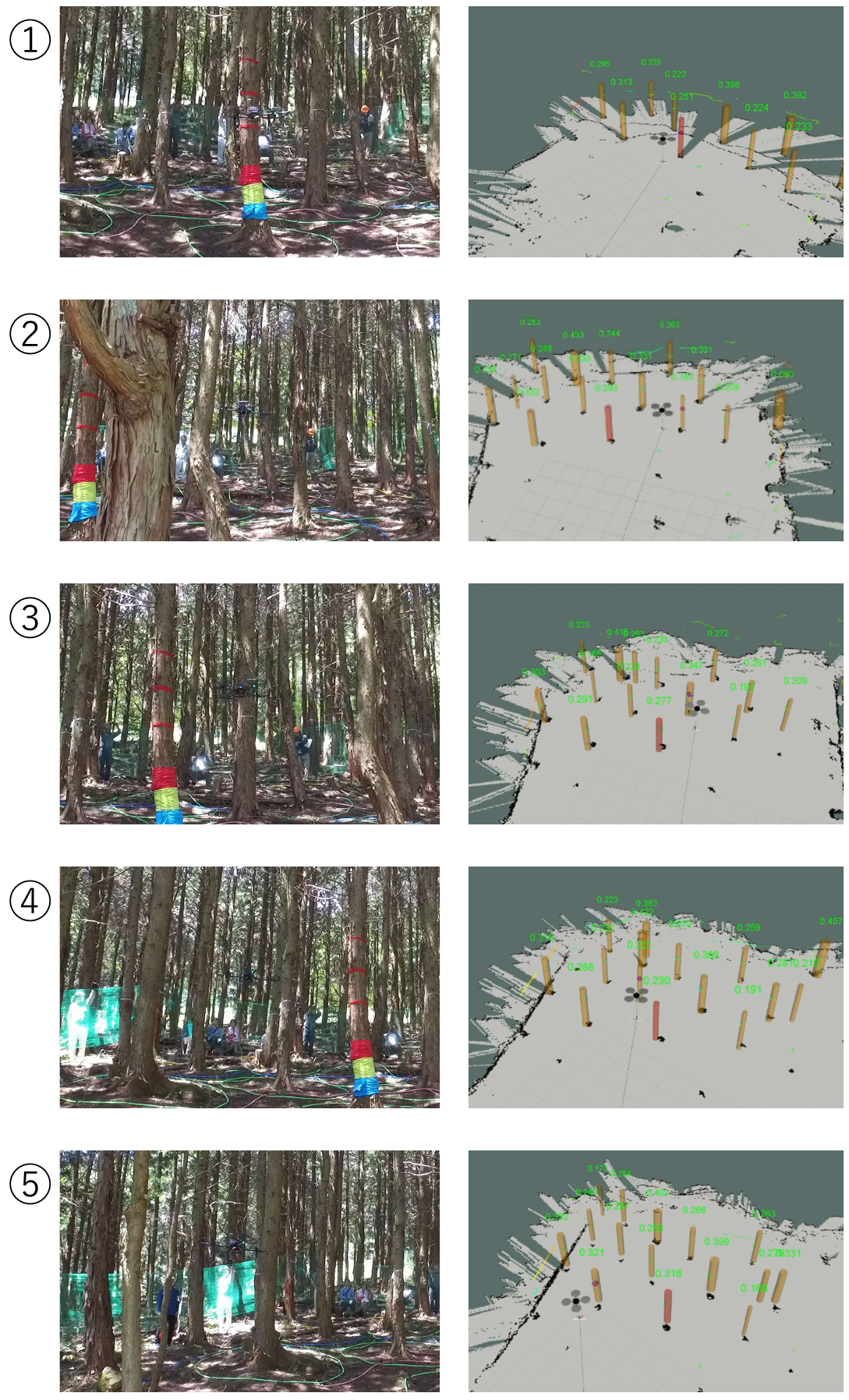}
\end{center}
\caption{Snapshots of the tree measurement experiment in forest drone robot competition at Yufuin. The aerial robot goes around five trees and measures diameter.
\label{figure:experiment_snapshot}}
\end{figure}
\begin{figure}[t]
\begin{center}
\includegraphics[bb=0 0 253 144, width=1.0\linewidth]{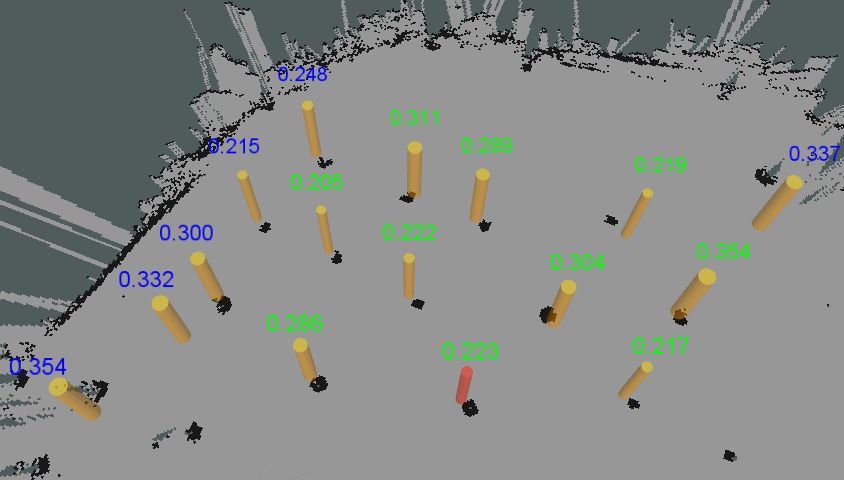}
\end{center}
\caption{Tree database and map generated in the flight experiment. The red tree is the labeled tree. The values above the trees are the estimated diameters. The values displayed in green are submitted and compared to the true diameters.
\label{figure:yufuin_final_result}}
\end{figure}

\figref{experiment_snapshot} shows the flight experiment in the third task. The aerial robot went around five trees including the labeled tree using narrow searching method. \figref{yufuin_final_result} shows the tree database and map generated in this experiment. The cylinders indicate measured trees. The red tree is the labeled tree. The values above the trees are the estimated diameters. Since the committee used only ten trees the diameter of which is displayed in green, we submitted the data of diameter of them. In \tabref{measurement_result}, we show the result of tree diameter measurement. The maximum error is about 0.08[m]. One of the reasons is the difference of the height used for measurement. When calculating true diameter, the measurer measures circumference of trees at the height of 1[m] and divides it by $\pi$. To avoid recognizing referees as trees, we set the hovering height of the aerial robot at 1.5 $\sim$ 2.0[m]. However, the average of the error is 0.034[m], which can be assumed as an allowable error. Therefore  we conclude that the effectiveness of the measurement method and navigation system of the aerial robot was demonstrated.

\begin{table}[th] 
 \caption{Result of tree diameter measurement.}
 \label{table:measurement_result}
 \centering
 \begin{tabular}{|l|l|l|l|}
  \hline
  tree number & estimated diameter[m] & true diameter[m] & error[m] \\ \hline 
  $1$ & 0.223 & 0.299 & 0.076 \\
  $2$ & 0.286 & 0.307 & 0.021 \\
  $3$ & 0.222 & 0.227 & 0.005 \\
  $4$ & 0.304 & 0.326 & 0.022 \\
	$5$ & 0.217 & 0.218 & 0.001 \\
	$6$ & 0.205 & 0.217 & 0.012 \\
	$7$ & 0.311 & 0.350 & 0.039 \\
	$8$ & 0.289 & 0.218 & 0.071 \\
	$9$ & 0.219 & 0.228 & 0.009 \\
	$10$ & 0.354 & 0.272 & 0.082 \\
  \hline
 \end{tabular}
\end{table}

%% file: src/conclusion.tex
\section{Conclusion}
In this paper, we investigated sensing and navigation system of aerial robot for measuring tree location and size in forest environment. We developed tree recognition and measurement method with a forest sensing device consisting of LRF and USB camera. We then proposed an algorithm to generate tree database. We described aerial robot navigation method including circular motion and searching method. Finally, we showed the effectiveness of the proposed method by an experiment. In the experiment, we showed the achievement of measuring tree location and size by an aerial robot flying automatically.
\par
In this work, we used range data from LRF for tree recognition and measurement. However, an object like cylinder can be judged as a tree by this method. Especially, it is dangerous if a human body is misrecognized and aerial robot approaches to the human. Therefore, for future work, we will develop image based tree recognition method with deep learning, for example. Moreover, in real forests, there are a lot of branches spread by trees. In our system, it is possible that aerial robot hit a branch because the branch is not recognized as an obstacle. Thus, we consider that branch recognition by RGBD sensor is necessary for safe flight.